\documentclass[preprint,review,12pt]{elsarticle}

\usepackage{booktabs} 
\usepackage{amsbsy,amsthm,amsmath,amssymb}
\usepackage{caption}
\usepackage{subcaption}
\usepackage{algorithm}
\usepackage{algorithmic}
\usepackage{xcolor}
\usepackage{multicol}
\usepackage{hyperref}
\hypersetup{
    colorlinks=true,
    linkcolor=black,
    filecolor=black,
    urlcolor=blue,
    citecolor=black,
}

\usepackage{stmaryrd}
\usepackage{enumitem}
\usepackage{bbm}
\usepackage{cleveref}
\usepackage{multirow}
\usepackage{subcaption}

\newcommand{\ba}{\boldsymbol{a}}
\newcommand{\bb}{\boldsymbol{b}}

\newcommand{\bk}{\boldsymbol{k}}

\newcommand{\bq}{\boldsymbol{q}}
\newcommand{\br}{\boldsymbol{r}}

\newcommand{\bv}{\boldsymbol{v}}

\newcommand{\bz}{\boldsymbol{z}}

\newcommand{\bW}{\boldsymbol{W}}
\newcommand{\bX}{\boldsymbol{X}}
\newcommand{\bY}{\boldsymbol{Y}}

\newcommand{\bbeta}{\boldsymbol{\beta}}

\newcommand{\bPhi}{\boldsymbol{\Phi}}

\newcommand{\bSigma}{\boldsymbol{\Sigma}}

\usepackage[english]{babel}

\renewcommand{\vec}[1]{{\boldsymbol{\mathbf{#1}}}}

\journal{Artificial Intelligence in Medicine}

\begin{document}

\begin{frontmatter}

\title{TransformerLSR: Attentive Joint Model of Longitudinal Data, Survival, and Recurrent Events with Concurrent Latent Structure}

\author[inst1]{Zhiyue Zhang}
\ead{zzhan179@jhu.edu}
\affiliation[inst1]{organization={Department of Applied Mathematics and Statistics,Johns Hopkins University},
            addressline={3100 Wyman Park Dr},
            city={Baltimore},
            postcode={21211},
            state={MD},
            country={USA}}

\author[inst1]{Yao Zhao}
\ead{yzhao136@jhu.edu}
\author[inst1,inst2]{Yanxun Xu\corref{cor1}}
\ead{yanxun.xu@jhu.edu}
\cortext[cor1]{Corresponding author.}
\affiliation[inst2]{organization={Division of Biostatistics and Bioinformatics, School of Medicine, Johns Hopkins University},
            addressline={733 N Broadway}, 
           city={Baltimore},
           postcode={21205},
            state={MD},
            country={USA}}

\begin{abstract}
In applications such as biomedical studies, epidemiology, and social sciences, recurrent events often co-occur with longitudinal measurements and a terminal event, such as death. Therefore, jointly modeling longitudinal measurements, recurrent events, and survival data while accounting for their dependencies is critical. While joint models for the three components exist in statistical literature, many of these approaches are limited by heavy parametric assumptions and scalability issues. Recently, incorporating deep learning techniques into joint modeling has shown promising results. However, current methods only address joint modeling of longitudinal measurements at regularly-spaced observation times and survival events, neglecting recurrent events. In this paper, we develop TransformerLSR, a flexible transformer-based deep modeling and inference framework to jointly model all three components simultaneously. TransformerLSR  integrates deep temporal point processes into the joint modeling framework, treating recurrent and terminal events as two competing processes dependent on past longitudinal measurements and recurrent event times. Additionally, TransformerLSR introduces a novel trajectory representation and model architecture to potentially incorporate {\it a priori} knowledge of known latent structures among concurrent longitudinal variables. We demonstrate the effectiveness and necessity of TransformerLSR through simulation studies and analyzing a real-world medical dataset on patients after kidney transplantation.

\end{abstract}


\end{frontmatter}


\section{Introduction}
\label{sec: intro}

Analyzing data from many biomedical studies often involve studying patients undergoing recurrent events, such as frequent clinic visits and hospital readmissions \cite{nevins2017understanding,fontanarosa2013revisiting}. At the occurrence of these events, patients' health-related measurements are recorded. Both the recurrent events and longitudinal data can be terminated by censoring or a failure event such as death. Our motivating application is data from patients after kidney transplantation, where routine outpatient visits are crucial for monitoring health and preventing graft rejection.  At each visit, important longitudinal measurements reflecting kidney function, such as creatinine levels, are recorded. The terminal event in this context could be either death or a return to dialysis. Understanding the relationship between patients' longitudinal measurements and the occurrence of clinic visits, as well as their combined impact on patient survival, is essential for making informed medical decisions. Therefore, it is important to jointly model longitudinal data, recurrent events, and survival to fully understand their interdependency.

Joint modeling of longitudinal and survival data have been widely studied in statistical literature \citep{wulfsohn1997joint,henderson2000joint,xu2001joint,tsiatis2004joint,crowther2013joint}.  Typically, these models employ submodels for each component, e.g., linear mixed-effects models for longitudinal data and Cox proportional hazards models for survival times, then link them through shared parameters such as random effects or frailty terms. For example, Hickey et al. proposed joineRML to jointly model multivariate longitudinal data and survival by assuming shared random effects \citep{hickey2018joinerml}. However, these models are limited by strict parametric assumptions in submodel choices and computational challenges, particularly with large datasets, due to the curse of dimensionality when integrating out a large number of random effects. To address these limitations, efforts have been made to leverage the advancements in deep neural network models for joint modeling. For example, recurrent neural networks have been utilized to encompass all historical data into a model's hidden state at the time of prediction, followed by prediction layers for longitudinal outcomes and survival risks \citep{lim2018disease}.  Similarly, MATCH-Net employed convolutional neural networks to identify temporal dependencies and conduct survival analysis based on longitudinal outcomes \citep{jarrett2019dynamic}. More recently, Lin et al. proposed TransformerJM by adapting the transformer architecture \citep{vaswani2017attention} for joint modeling and showed improved performance empowered by the attention mechanism \citep{lin2022deep}. Additionally, they explored a hybrid method named MFPCA-DS, which integrates MFPCA \citep{happ2018multivariate} for modeling multivariate longitudinal data and DeepSurv \citep{katzman2018deepsurv} for survival data analysis through neural networks. 

There has also been few work on jointly modeling of longitudinal data, recurrent events, and survival data. Most of them assumed shared random effects among the three submodels for longitudinal data, recurrent events, and survival, respectively \citep{liu2009joint,kim2012joint}. An exception is \citep{cai2017joint}, where the dependence among the three components was modeled by rescaling the time index. Despite these advancements, all these statistical methods heavily rely on specific parametric assumptions and entail complex inference procedures.  In the realm of deep learning, there has yet to be exploration into a joint model that encompasses longitudinal data, recurrent events, and survival data within a single, flexible framework. This gap exists due to the challenge of effectively handling continuous-time event stream modeling for recurrent events. Based on the classical neural network architecture designs, most mainstream deep joint methods can only suitably handle longitudinal and survival data observed at a fixed schedule with regular time intervals, e.g., daily, monthly. However, the time grids are distorted in the presence of recurrent events.

To address these challenges, we develop TransformerLSR, a novel and flexible continuous-time \textbf{transformer} for joint modeling of \textbf{l}ongitudinal, \textbf{s}urvival data, and \textbf{r}ecurrent events. TransformerLSR stands out with several key features. 
First, TransformerLSR models both recurrent events and survival events as competing temporal point processes with deep likelihood-based learning. This fully generative model suitably captures the stochastic nature of continuous-time recurrent events. To our best knowledge, TransformerLSR is the first deep joint model that integrates longitudinal data, survival data, and recurrent events in a single framework.
Second, the novel architecture of TransformerLSR allows to incorporate known clinical knowledge to aid inference and interpretability. By utilizing a trajectory representation which further separates multivariate longitudinal variables into single tokens, TransformerLSR is able to incorporate the deeper concurrent latent structure among the longitudinal variables beyond simple correlation. Additionally, this representation adeptly manages missing data by selectively marginalizing missing dimensions at each observation point, thereby maximizing the utility of available data. 
Third, the deep learning nature of TransformerLSR provides flexibility for modeling intensities of recurrent and survival events, setting it apart from deep survival models building on Cox regression \citep{katzman2018deepsurv,kvamme2019time}. 
Lastly, Leveraging the power of contemporary computing advancements, TransformerLSR's end-to-end deep model bypasses the need for complicated estimator derivations or complex sampling inference schemes often required by conventional statistical approaches.

The rest of this article is structured as follows. We present the proposed  TransformerLSR in \Cref{sec: methods}. In \Cref{sec: sim}, we evaluate TransformerLSR through a series of simulation studies, and compare against alternative approaches. In \Cref{sec: divat}, we apply TransformerLSR to a real-world kidney transplantation dataset. We conclude this work and discuss future directions in \Cref{sec: conclusion}.

\section{Methods}
\label{sec: methods}

 For each patient $i$, $i=1, \dots, I$, assume that we have baseline covariates denoted by $\bX_i$. Denote $E_i$ the death time and $C_i$ the administrative censoring time for patient $i$. Let $T_i=\min(E_i,C_i)$ and $\delta_i=\mathbb{I}_{(C_i\leq E_i )}$ denote the observed survival event time and the  censoring indicator, respectively. Assume that patient $i$ has $J_i$ recurrent events (e.g., clinic visit) at times $\{t_{i,j}\}_{j=1}^{J_i}$, where  $T_i\geq t_{i,J_i}$. At each time $t_{i,j}$, an $m$-dimensional longitudinal measurements   $\bY_{i,j}$ is recorded.  In the rest of this section, we suppress the patient index $i$  when the context is clear.

Denote the history of a patient up to the $j$th visit to be $\mathcal{H}_j=\{\overline{\bY}_j,\overline{t}_j,\bX\}$, where the longitudinal measurement history $\overline{\bY}_j=\{\bY_1,\dots,\bY_j\}$, and recurrent event times $\overline{t}_j=\{t_1,\dots,t_j\}$. Similarly, $\mathcal{H}(t)$ denotes the history recorded up to time $t$.  Our modeling framework is formulated as follows. We model  the longitudinal measurements  $\overline{\bY}_J$ as a stochastic process that exist and evolve in continuous time, and are observed at discrete sampling times $\overline{t}_J$. For the survival outcome, we assume that the survival event happens in an infinitesimally small interval $[t,t+dt)$ with probability $h(t)dt$, in which the hazard rate $h(t)$ is defined as $h(t)=\lim_{dt\rightarrow 0} \frac{\mathbb{P}\{t\leq E <t+dt|E\geq t,\mathcal{H}(t)\}}{dt}$. Therefore, the survival probability up to time $t$ is $S(t)=\exp\{-\int_0^t h(x)dx\}$. For modeling recurrent events, we employ a temporal point process and characterize event times using a conditional intensity function $\lambda(t)$, which
is the probability of observing an event in the time window $[t, t + dt)$ given the event history $\mathcal{H}(t)$. To effectively capture the interplay among longitudinal data, survival outcomes, and recurrent events and enable accurate predictions while accounting for their dependencies, we develop TransformerLSR building upon an encoder-decoder transformer architecture. This design provides a flexible and effective framework for joint modeling.

\subsection{Model Architecture}

\begin{figure}[ht!]
    \centering
    \includegraphics[width=1\linewidth]{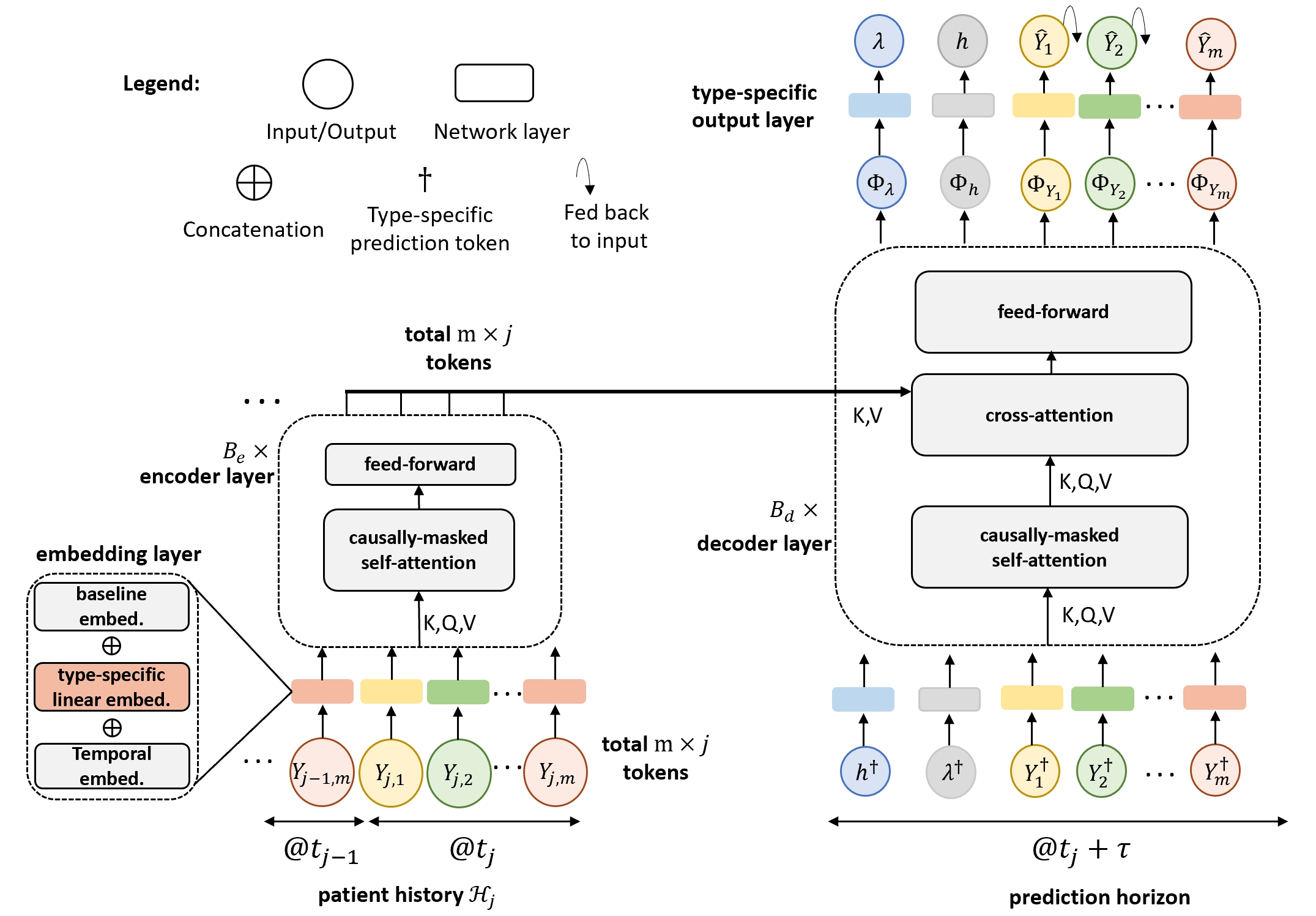}

        \caption{Architecture of TransformerLSR, layer normalization/residual connection omitted for clarity. The encoder processes the input patient history $\mathcal{H}_j$, and feeds to the encoder to output the event intensity $\lambda(t_j+\tau)$ and hazard $h(t_j+\tau)$ after lag time $\tau$. The longitudinal variables $Y_{1:m}(t_j+\tau)$ are predicted autoregressively, where each output $\widehat{Y}_u(t_j+\tau)$ is fed back to the decoder input for the prediction of subsequent variables. }
    \label{fig: arch}
\end{figure}

The overall architecture of TransformerLSR is shown in \Cref{fig: arch}. The backbone of TransformerLSR is an encoder-decoder transformer with the multi-head attention mechanism \citep{vaswani2017attention}, in which the encoder processes the input patient history $\mathcal{H}_j$ and feeds to the decoder to output the desired quantities $\widehat{\bY}(\tau|t_j)=\left(\widehat{Y}_u(\tau|t_j)\right )_{u=1}^m$, $\lambda(t_j+\tau)$, and $h(t_j+\tau)$, for any lag time $\tau$ into the future. In the rest of this subsection, we detail the architecture of our model, one component at a time.

\paragraph{Trajectory Representation and Causal Mask Incorporating Known Clinical Knowledge} 

Traditional deep learning models typically treat the $m$-dimensional longitudinal measurements $\bY$ as a multivariate variable,  assuming conditional independence among different dimensions given the history \citep{lim2018disease,lin2022deep}. However, recent successes in various applications, such as reinforcement learning \citep{janner2021offline} and natural language processing \citep{brown2020language}, have been observed with autoregressive generative models. These models depart from the traditional approach by assuming a sequential dependence among the various dimensions of variables. In this work, we align with this perspective and model longitudinal variables autoregressively. Specifically, we separate the multivariate variables and represent each dimension by its own token. As a result, for a given history $\mathcal{H}_j$, our trajectory representation of $\overline{\bY}_j$ takes the form $\left (Y_{1,1},\ldots,Y_{1,m},\ldots,Y_{j,1},\ldots,Y_{j,m}\right )$. 

 In many medical applications, known clinical knowledge may exist, such as instantaneous causal relationships among longitudinal variables. For example, in our motivating kidney transplantation application, the tacrolimus level in the blood may affect the creatinine level, but not vice versa. When such knowledge is available, we establish an order for concurrent tokens, ensuring that causative variables precede dependent variables in the sequence. A causal mask (lower triangular, as defined in \citep{vaswani2017attention}) is applied, allowing each token to attend to itself and all preceding tokens. 
  
For instance, consider concurrent variables $A$, $B$, and $C$ measured simultaneously. If known {\it a priori} that $A$ causes $B$, $B$ influences $C$, our trajectory representation becomes $(A, B, C)$. Here, $A$ attends only to itself, $B$ attends to $\{A, B\}$, and $C$ attends to $\{A,B,C\}$.

\paragraph{Encoder} The encoder processes the patient history $\mathcal{H}_j=\{\overline{\bY}_j,\overline{t}_j,\bX\}$, contextualizing each dimension of the longitudinal variable $\{{Y}_{j,u}\}_{(j,u)\in (1:J\times 1:m)}$ with prior history information. Given the history $\mathcal{H}_j$, we prepare the input sequence of the longitudinal variables $\overline{\bY}_j$ for the encoder using our trajectory representation $\left (Y_{1,1},\ldots,Y_{1,m},\ldots,Y_{j,1},\ldots,Y_{j,m}\right )$, spanning a total length of $m\times j$. To form the base-level input tokens $\{\bPhi_l^{(0)}\}_{l=1}^{m\times j}$ which incorporate temporal information, longitudinal data, and baseline covariates, we undertake several steps. First, we embed the baseline covariates, enabling the integration of patient-specific baseline information. Following this, we perform type-specific linear embeddings to the longitudinal measurements, employing distinct embedding layers for different types of measurements (e.g., $Y_{j,1}$ vs $Y_{j,2}$) to capture the unique characteristics of each measurement. 
Additionally, we adopt the sinusoidal temporal embedding, following prior works on continuous-time transformers \citep{zuo2020transformer,yang2022transformer, zhang2023continuous}, to allow the embeddings to vary continuously in time, ensuring that the temporal dynamics are fully represented. Specifically, 

\begin{equation*}
  \sin\left( \frac{t}{10000^{\frac{s}{d_{\text{time}}}}}   \right) \text{for even }s\text{,}
 \cos\left( \frac{t}{10000^{\frac{s-1}{d_{\text{time}}}}}   \right) \text{for odd }s,
\end{equation*}
\noindent where $s\in\{1,\ldots,d_{\text{time}}\}$, and $d_{\text{time}}$ is the temporal embedding dimension. 

In our adaptation of the transformer model \citep{vaswani2017attention}, instead of employing summation to integrate the three embeddings (baseline covariates, type-specific longitudinal measurements, and temporal information), we opt for concatenation. This choice facilitates direct access to and preserves the distinctiveness of the temporal information. 

Within each encoder layer $b$, where $b$ ranges from $1$ to $B_e$ (the total number of encoder layers), we compute the key, query, and value $\bk_l^{(b)},\bq_l^{(b)},\bv_l^{(b)} \in \mathbb{R}^{d_{\text{model}}}$, with $d_{\text{model}}$ denoting the model dimension. They are obtained by $\bk_l^{(b)}=K ^{(b)}(\bPhi_l^{(b-1)}),\quad \bq_l^{(b)}=Q ^{(b)}(\bPhi_l^{(b-1)}), \quad \bv_l^{(b)}=V ^{(b)}(\bPhi_l^{(b-1)}),$

where $K^{(b)},Q^{(b)},V^{(b)}$ are the linear transformations for keys, queries, and values, respectively. Utilizing the causal mask described previously, each $\bPhi_l^{(b-1)}$ attends only to its preceding tokens and itself $\{\bPhi_z^{(b-1)}\}_{z=1}^{l}$, with unnormalized attention weight paid to element $\bPhi_z^{(b-1)}$ given by $\alpha(\bPhi_{l}^{(b-1)},\bPhi_z^{(b-1)})=\bq_l^{(b)}\cdot \bk_z^{(b)}$. Then, we compute 
$\sum _{z=1}^{l}\texttt{softmax}\big(\{\alpha(\bPhi_l^{(b-1)},\bPhi_{z'}^{(b-1)})\}_{z'=1}^{l}\big)_z  \cdot \bv_z^{(b)}$,
followed by dropout \citep{srivastava2014dropout}, residual connection \citep{he2016deep}, layer normalization \citep{ba2016layer}, and feed-forward network (with ReLU activation \citep{ramachandran2017searching}) to obtain the layer-level output $\bPhi_l^{(b)}$. The final layer outputs $\{\bPhi_l^{(B_e)}\}_{l=1}^{m\times j}$, enriched with contextual information from the patient’s history, serves as the input to the decoder for further processing, acting as the keys and values.

\paragraph{Decoder} The decoder takes the processed input history $\mathcal{H}_j$ from the encoder, and outputs $\widehat{\bY}(\tau|t_j)=\left(\widehat{Y}_u(\tau|t_j)\right )_{u=1}^m$, the conditional intensity function 
 $\lambda(t_j+\tau)$, and the hazard function $h(t_j+\tau)$, for any given lag time $\tau>0$. 
 To achieve this, the decoder employs type-specific prediction tokens for each output quantity, which are embedded using the same methodology as the encoder. These prediction tokens then participate in the attention mechanism, where, in adherence to the causal mask, they attend to themselves and the tokens preceding them to grasp the current time and concurrent information. Additionally, they attend to all input tokens from the encoder to incorporate historical context.

The attention computation is followed again by dropout, residual connection, layer normalization, and feed-forward network. This sequence is repeated across decoder layers $b$, for $b$ ranging from 1 to $B_d$, the total number of decoder layers. The output from the final decoder layer, denoted as $\bPhi:=\bPhi^{(B_d)}$, is then directed through task-specific prediction layers. 
The conditional intensity function for a future event is calculated as $\lambda(t+\tau) = \text{Softplus}(\bW_{\lambda}\bPhi_{\lambda}+\ba_{\lambda})$, and 
the hazard function for the survival outcome is determined by $h(t+\tau) = \text{Softplus}(\bW_{h}\bPhi_{h}+\ba_{h})$. The predicted value for each longitudinal measurement $u$, given a lag time $\tau$ from the current time $t_i$, is given by $\widehat{Y}_u(\tau|t_i) = \bW_{Y_u} \bPhi_{Y_u}+\ba_{Y_u}$, for each $u \in \{1,\ldots,m\}$.

Here, the weight $\bW_{(\cdot)}$ and bias $\ba_{(\cdot)}$ specific to each output  are parameters that the model learns through training. The softplus operation is applied for the event intensity and hazard rate to ensure they are greater than or equal to zero. For longitudinal variables, we predict in an autoregressive fashion. Therefore, the predicted value of each longitudinal variable is iteratively fed back into the decoder input, replacing its respective prediction token. 

\par In scenarios where the timings of recurrent events may differ from those of longitudinal measurements, TransformerLSR offers a flexible adaptation by processing two distinct sets of input times within the encoder. Consequently, it can simultaneously generate predictions for event intensities and longitudinal measurements at their respective times in the decoder. This adaptability ensures that the model remains applicable across a wide range of real-world settings where data collection schedules for different types of events may not align. 

Moreover, due to our sequential trajectory representation, our model architecture can handle asynchronous missing data effectively. Unlike approaches that necessitate discarding an entire multivariate observation due to partial missingness \citep{jarrett2019dynamic}, our model employs selective masking. This technique involves masking only the missing dimensions within an observation, preserving the non-missing components for analysis. Such a strategy significantly enhances the model's ability to leverage the available data, minimizing the loss of valuable information due to incomplete data points. We provide a more detailed discussion on asynchronous missing data handling in \ref{sec: append_missing}.

\subsection{Training}
The  training loss for our model is the sum of three components: $\mathcal{L} = \mathcal{L}_{\vec{Y}}+\mathcal{L}_{\lambda} + \mathcal{L}_h$, where $\mathcal{L}_{\vec{Y}}$, $\mathcal{L}_{\lambda}$, and $\mathcal{L}_h$ are longitudinal variable loss, recurrent event loss, and survival event loss, respectively. For the longitudinal variables, we employ the mean squared error (MSE) to quantify the discrepancy between predicted and observed values: $ \mathcal{L}_{\vec{Y}} = \frac{1}{mJ}\sum_{u=1}^m\sum_{j=1}^J (\widehat{Y}_{j,u}-Y_{j,u})^2.$
 
For the recurrent event intensity and survival hazard, we employ maximum-likelihood training, with the losses formulated as the negative log-likelihoods of the respective processes, that is, $\mathcal{L}_{\lambda} = -l_{\lambda}$, $\mathcal{L}_h = -l_h$. For recurrent events over the time interval $[0,T)$, the log-likelihood is that of a general temporal point process:  
\begin{equation}
    l_{\lambda} = \underbrace{\sum_{j=1}^{J} \log \lambda(t_j)}_{\text{event log-likelihood}} \qquad - \underbrace{\int_{0}^T \lambda(t) dt}_{:=\Lambda, \text{ non-event log-likelihood}} \! \! \! \! \!\!.
    \label{eq: event_loglik}
\end{equation}

We provide a proof to \Cref{eq: event_loglik} in \ref{sec: append_like} for completeness. Similarly, subject to right censoring and having at most one survival event, the log-likelihood for the survival data is 
\begin{equation}
l_h = (1-\delta) \log h(T) - \underbrace{\int_0^T h(t)dt}_{:=\zeta}.
\label{eq: surv_loglik}
\end{equation}
While it is straightforward to compute the \emph{event log-likelihood} in \Cref{eq: event_loglik,eq: surv_loglik}, the \emph{non-event log-likelihood} therein involves integral computations and therefore poses a computational challenge. We tackle this with the Monte Carlo trick: to estimate the integral $\Lambda$, we sample a total of $N$ samples of $\lambda(t_{j,n})$ where $t_{j,n}\sim \mathcal{U}(t_j,t_{j+1})$ for each $j$ (with $t_{J+1}:=T$ in the case $T>t_{J})$, resulting in an unbiased estimate $\widehat{\Lambda}_{\text{MC}}$ for $\Lambda$: $\widehat{\Lambda}_{\text{MC}} = \sum_{j=1}^J (t_{j+1}-t_{j})\left(\frac{1}{N}\sum_{n=1}^N \lambda (t_{j,n})\right).$

To compute each $\lambda(t_{j,n})$, we embed an extra prediction token $\lambda^{\dagger}$ at time $t_{j,n}$, which only has access to history $\mathcal{H}_j$ up to time $t_j$. We obtain $\widehat{\zeta}_{\text{MC}}$  analogously with samples of $h(t_{j,n})$.

\subsection{Inference} 
For longitudinal outcomes, our model utilizes the history $\mathcal{H}_j$ to generate point estimates $\widehat{\bY}(\tau|t_j)$, allowing for predictions into the future at any given lag time \(\tau\). This capability facilitates dynamic forecasting based on the most recent data available.

In estimating survival probabilities conditioned on a lag time \(\tau\), we employ a Monte Carlo approach to draw samples \(\{h(t_{j,n})\}\) where each \(t_{j,n}\) follows a uniform distribution \(\mathcal{U}(t_j,t_j+\tau)\). This sampling strategy aids in computing the conditional survival probability as \(\mathbb{P}\{E\geq t_j+\tau |E\geq t_j,\mathcal{H}_j\} = \exp \{-\int_{t_j}^{t_j+\tau} h(x)dx\}\), effectively translating the hazard function samples into a probability measure for the survival beyond the time \(t_j+\tau\), given survival up to time \(t_j\) and the history up to that point.

For recurrent events, the probability of an event not occurring within a lag time \(\tau\) from the last event, \(\mathbb{P}\{t_{j+1}\geq t_{j} + \tau|\mathcal{H}_j\}\), is similarly derived using Monte Carlo samples of the intensity function \(\{\lambda(t_{j,n})\}\). This probabilistic framework not only allows for the estimation of event likelihoods but also sets the stage for more detailed event time sampling.

To sample future event times \(t_{j+1}\) based on these probabilities, our model adopts the thinning algorithm \citep{lewis1979simulation}, circumventing the need for repetitive computation of Monte Carlo integrals as required by inverse transform sampling. The thinning algorithm offers an efficient means to simulate event times from the conditional intensity functions, thereby enhancing the practicality and scalability of our predictive modeling. The details of the thinning algorithm are presented in \ref{sec: append_thinning}.

\section{Simulation study}
\label{sec: sim}

We assessed the proposed TransformerLSR by conducting simulation studies to evaluate its performance in inference tasks such as predicting longitudinal measurements, fitting survival functions, and estimating recurrent event intensities. As TransformerLSR represents the first effort in the literature to jointly model longitudinal measurements, recurrent events, and survival time, there currently do not exist any directly comparable methods in the literature. To showcase the superior performance of the proposed TransformerLSR, we examined four alternative methods: TransformerJM, MFPCA-DS, MATCH-Net, and joineRML. It is worth noting that all these alternative approaches are limited to joint modeling of longitudinal and survival data, lacking the functionality to incorporate recurrent events. Moreover, the comparative analysis required some adjustments due to the limitations of the alternative methods, particularly regarding their handling of missing data and irregular observation times. As TransformerJM, MFPCA-DS, and  joineRML do not directly handle missing data, we made the following minimal but necessary modifications: for TransformerJM, we masked off the multivariate longitudinal variables if any dimension was missing; for MFPCA-DS and joineRML, we performed the last observation carried forward imputation for the missing data. Furthermore, since MFPCA-DS and MATCH-Net cannot handle irregular observation times for longitudinal variables (since recurrent events happen in continuous time), we input the discrete observation indices in their methods, rather than the absolute observation times.

\subsection{Simulation setup}

Our simulation settings were designed to mimic the motivating kidney transplantation application. Assume that there were $I=1000$ patients, and each patient $i$ had three baseline covariates $\bX_i=(X_{i,1}$, $X_{i,2}$, $X_{i,3})$, where $X_{i,1}$ and $X_{i,2}$ were independently generated from a standard normal distribution $\mathcal{N}(0,1)$, and $X_{i,3} \sim \text{Bernoulli}(0.4)$, $i=1, \dots, I$. Suppose there were three longitudinal variables measured at clinic visit time $t_{i,j}$: $Y_{i,j,1}$, $Y_{i,j,2}$, and $Y_{i,j,3}$, which emulated the trough level of tacrolimus, creatinine level, and assigned tacrolimus dosage, respectively, for $j=1, \dots, J_i$.  Assume that the creatinine level $Y_{i,j,2}$ was influenced by the trough level of creatinine $Y_{i,j,1}$. Based on $Y_{i,j,1}$ and $Y_{i,j,2}$, the physician prescribed the dosage $Y_{i,j,3}$, which the patient was advised to follow until the next visit at $t_{i, j+1}$. Note that $Y_{i, j,u}=Y_{i, u}(t_{i,j})$, $u=1, 2, 3$. The longitudinal variables were generated as follows.  For tacrolimus trough levels, $Y_{i,1}(t)=Y_{i,1}^*(t)+\epsilon_{i,1}(t)$, where $\epsilon_{i,1}(t) \overset{\mathrm{iid}}{\sim}\mathcal{N}(0,0.1^2)$. The mean process $Y_{i,1}^*(t)$ was modeled by a linear mixed-effects model: $Y_{i,1}^*(t) = \bz_{i,1}(t)\beta_{l,1}+\br_{i,1}(t)\bb_{i,1}$, where the fixed effect covariates $\bz_{i,1}(t)=(1,Y_{i,3}(t),\bX_i,t)$, the random effect covariates $\br_{i,1}(t)=(1,Y_{i,3}(t),t)$. Here $Y_{i,3}(t)$ at any time $t$ refers to the dosage determined at the most recent clinic visit, meaning $Y_{i,3}(t)=Y_{i,j,3}$ for $t\in (t_{i,j}, t_{i, j+1}]$. Similarly, for creatinine levels, $Y_{i,2}(t)=Y_{i,2}^*(t)+\epsilon_{i,2}(t)$, where  $\epsilon_{i,2}(t) \overset{\mathrm{iid}}{\sim}\mathcal{N}(0,0.1^2)$. The mean process $Y_{i,2}^*(t)$ was also modeled by a linear mixed-effects model: $Y_{i,2}^*(t) = \bz_{i,2}(t)\beta_{l,2}+\br_{i,2}(t)\bb_{i,2}$, where $\bz_{i,2}(t)=(1,Y_{i,3}(t),\bX_i,Y_{i,1}(t),t)$ and $\br_{i,2}(t)=(1,Y_{i,3}(t),t)$. We set $\beta_{l,1}=(2.0,0.3,0.1,0.6, 0.2, -1\times 10^{-4})$, $\beta_{l,2}=(3.3,0.1,0.3,0.4, 0.25,1.0, -1\times 10^{-4})$, and sampled $\bb_{i,1},\bb_{i,2}$ independently from multivariate normal $\mathcal{N}(\mathbf{0},\bSigma_b)$, with $\bSigma_b = \text{diag}(0.2^2,0.07^2,1\times 10^{-8})$. At each clinic visit, the dosage was generated by $Y_{i,j,3}=(1,Y_{i,j,2},\bX_i)\bbeta_{l,3}+\epsilon_{i,j,3}$, where $\epsilon_{i,j,3} \overset{\mathrm{iid}}{\sim} \mathcal{N}(0,0.3^2)$, and $\bbeta_{l,3}=(1,0.2,0.15,0.2,0.15)$. Consequently, the dosage $Y_{i,3}(t)$ was a piecewise constant variable, whereas $Y_{i,1}(t)$ and $Y_{i,2}(t)$ varied continuously in time.

The recurrent event times $\{t_{i,j}\}_{j=1}^{J_i}$ were sampled from an inhomogeneous point process. The intensity function $\lambda(t)$ was set to depend on the longitudinal measurement of creatinine levels: $\lambda(t) = 3 \times \exp\left(-\left(Y_{i,2}^*(t) + 1.5\right)\right) t^{0.25}$.  For the survival times, we sampled with a Weibull proportional hazards model incorporating longitudinal dependencies: : $h_i(t) = \exp \big(-(1+ Y^*_{i,2}(t)+0.9 Y_{i,3}(t) ) \big)\omega t ^{\omega -1},$ where $\omega=1.25$.  This setup allowed the hazard rate for survival to vary as a function of both the creatinine levels and tacrolimus dosage over time. 

In our study, we explored the impact of different censoring rates on model performance by sampling censoring times $C_i$ from three distinct distributions: $\mathcal{N}(15000,100^2)$, $\text{Weibull}(2,8000)$, and $\mathcal{N}(1000,100^2)$. For each censoring distribution, we conducted $100$ repeated simulations. Averaging over $100$ simulations, the first censoring distribution resulted in a censoring rate of $2\%$, the second $13\%$, and the third $59\%$. To mimic the real-world missing data scenario, after sampling each dataset, we randomly masked off $25\%$ of $Y_1$, $15\%$ of $Y_2$, and $3\%$ of $Y_3$. For each dataset, we randomly chose $60\%$ of the patient trajectories as the training data, $20\%$ as the validation data, and the remaining $20\%$ as the evaluation data which our model had no access to during the training process. We set the trajectory representation to be $(Y_{j,1},Y_{j,2},Y_{j,3})$ for the concurrent variables. We set the number of encoder layers $B_e = 2$, and number of decoder layers $B_d = 3$, model dimension $d_{\text{model}}=64$, and used a dropout value of $0.2$. We used the Adam optimizer \citep{kingma2014adam}, with a learning rate of $1E-4$, and trained the model for $20$ epochs. When comparing our TransformerLSR model with TransformerJM, which also utilizes a transformer architecture, we aligned hyperparameters where feasible to ensure a fair comparison.  For other comparative methods, we adhered to the default settings provided by their respective packages.

 \subsection{Simulation results}

\par We first report the performance of TransformerLSR in predicting longitudinal measurements using the root mean square error (RMSE) as the evaluation metric. The results of TransformerLSR and all alternative methods  are summarized  in \Cref{tab: long_sim_missing}, where we report the mean results over $100$ repeated simulations (± standard deviation). MATCH-Net is left out as it does not model longitudinal outcomes. Across the three censoring settings, the proposed TransformerLSR outperformed the alternative methods. Specifically, for $Y_1$ prediction, TransformerLSR achieved a result comparable to that given by MFPCA when the censoring rate was 2\%, and outperformed all alternatives in the other two censoring rate settings. Notably, for $Y_2$ predictions, TransformerLSR significantly outperformed all alternatives in all settings, and for $Y_3$, TransformerLSR was on par with joineRML and both outperformed all the other methods. This showcases the effectiveness of using the concurrent latent structure for longitudinal measurements prediction, as TransformerLSR models the variables autoregressively, while all alternatives simply treat the longitudinal variables as multivariate variables.

\begin{table}[ht]
\caption{Longitudinal variables prediction error (RMSE) under three different censoring distributions.}
\label{tab: long_sim_missing}
\begin{center}
\footnotesize
\begin{tabular}{cc|cccc}
\hline
\multicolumn{2}{l|}{Censoring} & TransformerLSR & TransformerJM & MFPCA-DS & joineRML \\
\hline 
\multirow{3}{*}{2\%}
& $Y_1$ & 0.38±0.04 & 0.44±0.05 & 0.33±0.01 & 0.36±0.02 \\
& $Y_2$ & 0.44±0.04 & 0.61±0.07 & 0.55±0.02 & 0.54±0.03 \\
& $Y_3$ & 0.32±0.01 & 0.35±0.02 & 0.39±0.01 & 0.32±0.01 \\
\hline
\multirow{3}{*}{13\%}
& $Y_1$ & 0.32±0.03 & 0.40±0.04 & 0.34±0.01 & 0.35±0.02 \\
& $Y_2$ & 0.40±0.03 & 0.58±0.05 & 0.58±0.02 & 0.53±0.03 \\
& $Y_3$ & 0.32±0.01 & 0.36±0.02 & 0.39±0.01 & 0.32±0.01 \\
\hline
\multirow{3}{*}{59\%}
& $Y_1$ & 0.26±0.02 & 0.36±0.04 & 0.45±0.03 & 0.35±0.02 \\
& $Y_2$ & 0.36±0.02 & 0.54±0.04 & 0.79±0.05 & 0.56±0.03 \\
& $Y_3$ & 0.32±0.01 & 0.37±0.04 & 0.43±0.01 & 0.33±0.02 \\
\hline
\end{tabular}
\end{center}
\end{table}

\par We next examined the performance of TransformerLSR in modeling the survival events. Since TransformerLSR estimates the underlying hazard function $h(t)$, a straightforward evaluation of survival fitting is to compare the fitted log-likelihood of each patient's survival with the ground truth. The RMSE of log-likelihood was 0.67±0.09 for the $2\%$ censoring rate setting, 0.59±0.08 for the $13\%$ censoring rate setting, and 0.42±0.05 for the $59\%$ censoring rate setting. Note that we only report the RMSE of TransformerLSR since no alternative method outputs the continuous-time hazards. For context, the ground truth of log hazard function values ranged from $-4$ to $-12$. Evidently, TransformerLSR achieved good fitting and generalization ability for all three settings.

\par In order to compare against the alternatives, we employed a landmark analysis approach by selecting a landmark time $t_{\text{LT}}$ chosen as the $10\%$ quantile of death times in the training dataset. This strategy allowed us to assess the \emph{conditional} survival function $S(t|\mathcal{H}(t_{\text{LT}}), E\geq t_{\text{LT}})$ for each method, focusing on those patients in the evaluation set who survived past the landmark time.  We considered the integrated Brier score as the evaluation metric calculated by $\text{iBS}=\frac{1}{t_{\max}-t_{
\min}}\int_{t_{\min}}^{t_{\max}}\text{BS}(t) dt$, where $\text{BS}(t)$ is the Brier score \citep{rufibach2010use,blanche2015quantifying} evaluated at time $t$ and weighted by the censoring probability at $t$. To numerically approximate the integration required for iBS, we calculated the Brier score at six evenly spaced time points between the $10\%$ and $90\%$ quantiles of death times, employing the trapezoidal rule for integration. \Cref{tab: brier_sim_missing} summarizes the iBS results of all algorithms.

\begin{table}[ht]
\caption{Integrated Brier score (iBS) under three different censoring distributions.}
\label{tab: brier_sim_missing}
\begin{center}
\scriptsize 
\begin{tabular}{c|ccccc}
\hline
 Censoring &   TransformerLSR & TransformerJM & MFPCA-DS & MATCH-Net & joineRML \\
\hline 
$2\%$  & 0.16±0.02 & 0.52±0.03 & 0.20±0.03 & 0.21±0.06 & 0.19±0.02 \\
$13\%$  & 0.16±0.01 & 0.47±0.03 & 0.21±0.03 & 0.22±0.05 & 0.18±0.03 \\
$59\%$  & 0.14±0.02 & 0.21±0.02 & 0.16±0.02 & 0.25±0.02 & 0.14±0.02 \\
\hline
\end{tabular}
\end{center}
\end{table}

\par Across the three settings, TransformerLSR performed on par with MFPCA-DS and joineRML, and these hazard-based methods significantly outperformed TransformerJM and MATCH-Net. In particular, TransformerLSR, FMPCA-DS, joineRML, and MATCH-Net were all minimally impacted by the change of the censoring distribution, whereas TransformerJM changed dramatically in performance as the censoring rate changed.

\par
\begin{figure}
  \centering
    \includegraphics[width=\linewidth]{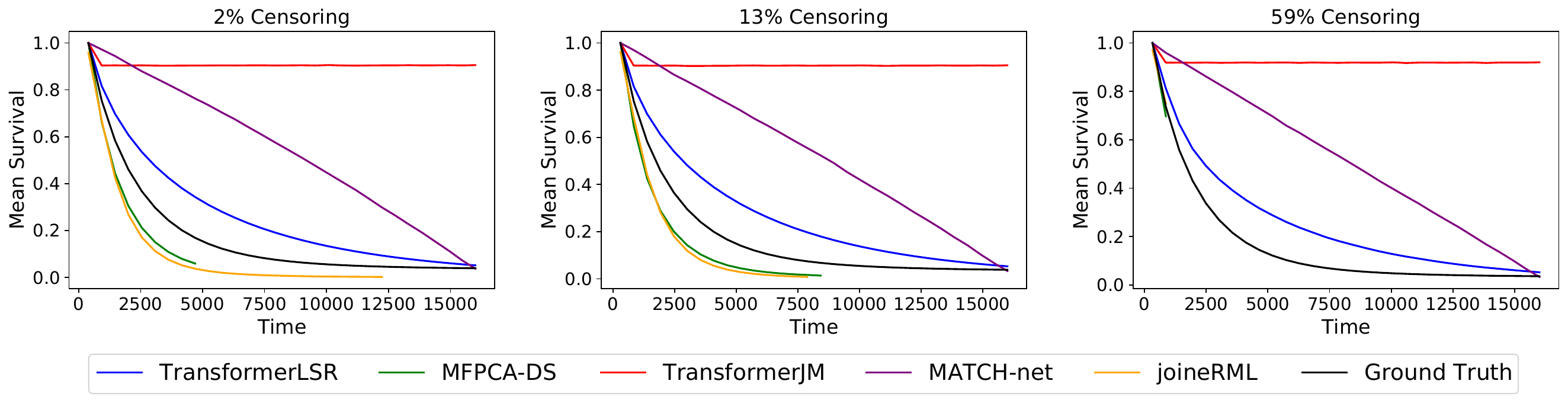}
\caption{Mean survival curves under three different censoring distributions.}
\label{fig: sim_survival}
\end{figure}

 To get a more thorough understanding of the behavior of each method beyond the numerically approximated iBS, we computed the conditional survival functions from the landmark time $t_{\text{LT}}$ to $t=16000$, a time slightly larger than the largest observed death time in all datasets. This analysis aimed to compare the mean survival functions predicted by each method against the ground truth across different censoring scenarios, as illustrated in \Cref{fig: sim_survival}. We again observed that TransformerLSR closely aligned with the ground truth in the $2\%$ censoring rate setting. Further, due to the nature of the Breslow estimator \citep{breslow1975analysis} used by DeepSurv and joineRML, these two methods were not capable of computing the survival function beyond the maximum death time in the training data (thus incomplete survival curves in all three settings in \Cref{fig: sim_survival}), while our intensity-based method, being fully generative, could extrapolate for all times. It is also worth noting that TransformerJM, whose survival modeling choice is conditional survival probability instead of a standard hazard-based formulation, exhibited profiles far off from the truth. Our conjecture for the poor performance of TransformerJM is that, since its survival formulation was originally made for the discrete time setting, the adopted binary survival loss was inaccurately influenced by the abundance of follow-up visits without recorded death events. Such imbalance likely skewed TransformerJM's predictions towards uniformly high survival probabilities, failing to differentiate effectively between varying levels of risk among patients.

\begin{figure}[ht]
    \centering
    \begin{subfigure}{.33\textwidth}
        \centering
        \includegraphics[width=\linewidth]{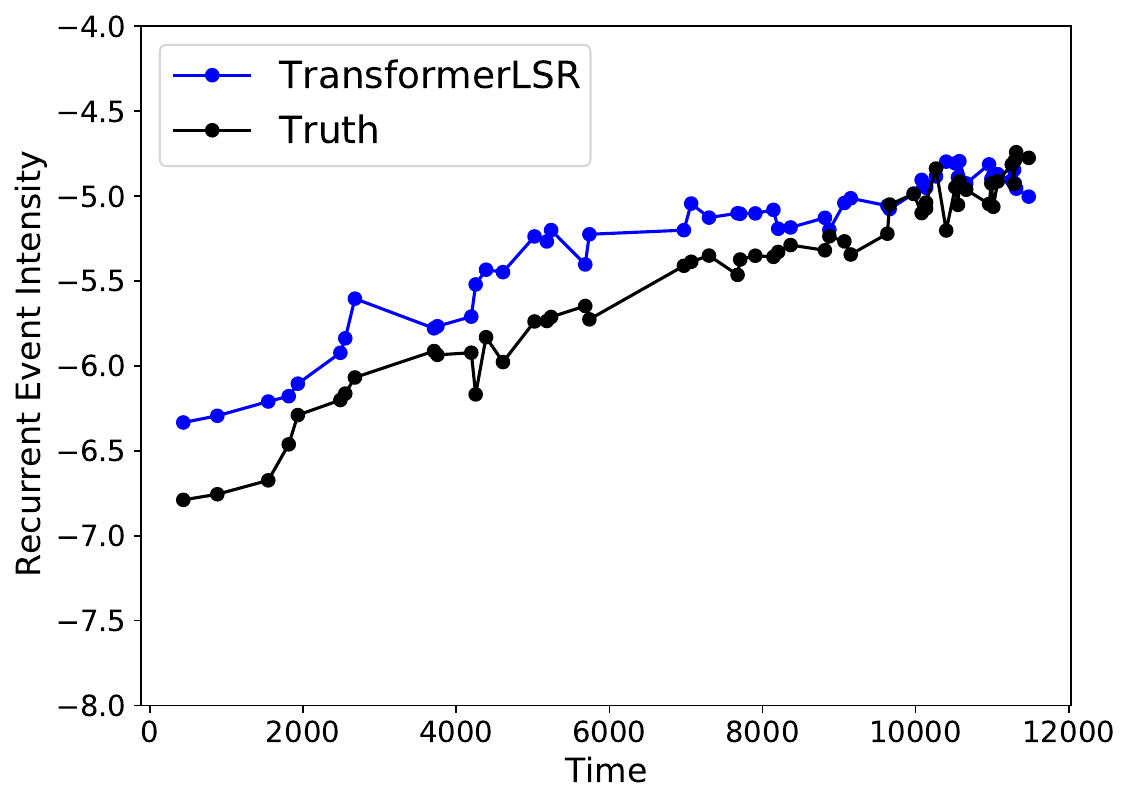}
        \caption{2\% Censoring.}
    \end{subfigure}%
    \begin{subfigure}{.33\textwidth}
        \centering
        \includegraphics[width=\linewidth]{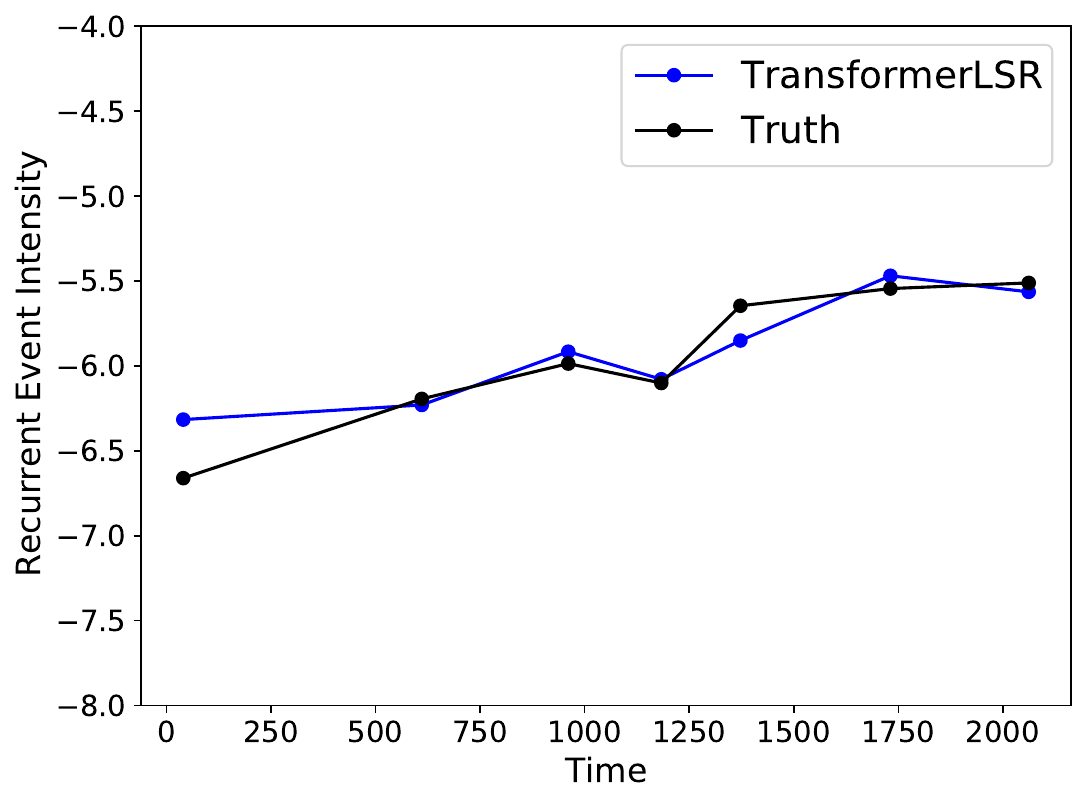}
        \caption{13\% Censoring.}
    \end{subfigure}
    \begin{subfigure}{.33\textwidth}
        \centering
        \includegraphics[width=\linewidth]{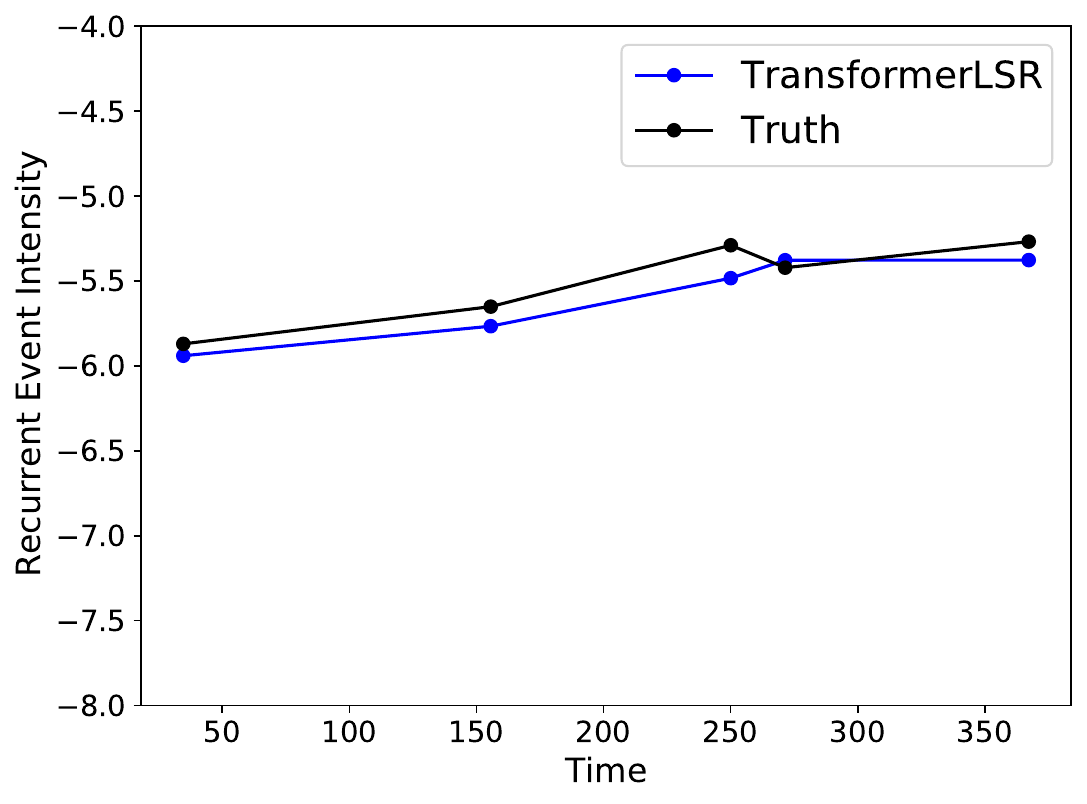}
        \caption{59\% Censoring.}
    \end{subfigure}
    \caption{Sample log recurrent event intensity function compared with the ground truth under the three settings.}
    \label{fig: sim_inten}
\end{figure}

\par Lastly, we turn our attention to recurrent event modeling. As the proposed TransformerLSR is the only approach that models continuous-time recurrent events, we only reported the results under TransformerLSR by computing the recurrent events log-likelihood in the evaluation data and comparing to the ground truth.  The RMSE of log-likelihood was 1.06±0.37 for the $2\%$ censoring rate setting, 0.84±0.20 for the $13\%$ censoring rate setting, and 0.50±0.08 for the $59\%$ setting. For context, the ground truth log event intensity function values ranged from $-3$ to $-9$. Our deep temporal point process model again achieved good fitting and generalization ability in all three settings. To visually assess the performance, \Cref{fig: sim_inten} plots a sample trajectory from the first simulation setting and compare that with the simulated true intensity function.

\section{Application to DIVAT}

\label{sec: divat}

Kidney transplantation is the primary therapy
for patients with end-stage kidney diseases \citep{arshad2019comparison}. French computerized and validated data in transplantation (DIVAT, \texttt{www.divat.fr}) is a data repository storing medical records of kidney transplantations conducted in various French hospitals, such as Nantes and Paris Necker. These medical records span from the transplantation date to the occurrence of graft failure, defined as either a return to dialysis  or patient death. During each clinic visit, crucial measurements reflecting kidney function and drug exposure—such as creatinine levels and trough levels of immunosuppressive medications like tacrolimus—are longitudinally recorded. Based on these measurements, clinicians determine tacrolimus dosages and schedule subsequent clinic visitations.

\par  We extracted 1443 adult patient records from Nantes involving kidney transplantations conducted between January 1, 2000, and December 31, 2014. Among the 1443 patients, we kept those with both baseline and follow-up records, and excluded patients with missing data in their baseline records. To mitigate the impact of short-term post-operative variations, we considered only follow-up records commencing six months after the transplantation. This approach also involved removing any follow-up visits recorded within a 5-day span of the previous visit, to avoid redundancy and potential data noise. Additionally, we excluded patients whose follow-up comprised a single visit, as this would not provide sufficient data for meaningful longitudinal analysis.
Finally, we excluded patients for whom any key variable -- assigned tacrolimus dosage, trough level of tacrolimus, or creatinine level -- was entirely missing across all follow-up visits.
Our filtering criterion yielded a total of $I=1238$ patient records. For each patient, we included a total of 23 baseline covariates, such as donor and recipient age and gender, occurrence of delayed graft function (defined as the use of dialysis within the first week post-transplant), and recipient body mass index (BMI). Additionally, we considered three longitudinal variables: creatinine levels, trough levels of tacrolimus, and prescribed dosages of tacrolimus at each clinic visit. To ensure numerical stability, all three longitudinal variables were transformed to the log scale. Sample trajectories of log-scale creatinine levels and tacrolimus dosages are provided in \ref{sec: append_divat}. We treated clinic visits as recurrent events and graft failure time as the survival event time. The survival censoring rate was $74.5\%$.

\par We applied the proposed TransformerLSR to the DIVAT data. In our model architecture, we incorporated the knowledge that the trough level of tacrolimus influences the creatinine level, and both variables affect the assigned tacrolimus dosage. As a result, we structured the trajectory representation with the order being (trough level of tacrolimus, creatinine level, tacrolimus dosage). For training and evaluation, we randomly selected $70\%$ of total patient trajectories for training, reserving the remaining $30\%$ as held-out data. We configured the model with $2$ encoder layers ($B_e$), $3$ decoder layers ($B_d$), and a model dimension of $d_{\text{model}}=64$, while applying a dropout value of $0.2$. Optimization was performed using the Adam optimizer with a learning rate set to $1E-4$, and the model was trained for $50$ epochs. The training plot for the training loss as a function of epochs is given in \ref{sec: append_divat} to show the sufficiency of $50$ epochs.

\par We first examined the performance of TransformerLSR in predicting longitudinal variables by comparing the one-step ahead prediction values to the observed values on the held-out data. On the log scale, the RMSE for trough level of  tacrolimus was 0.31, for creatinine level was 0.20, and for assigned tacrolimus dosage was 0.22. These errors were reasonably low, considering the dataset mean of 2.10 for the trough level of tacrolimus, 4.95 for creatinine level, and 1.71 for tacrolimus dosage.

\par
\begin{figure}
    \centering
    \begin{subfigure}{\textwidth}
        \centering
        \includegraphics[width=\linewidth]{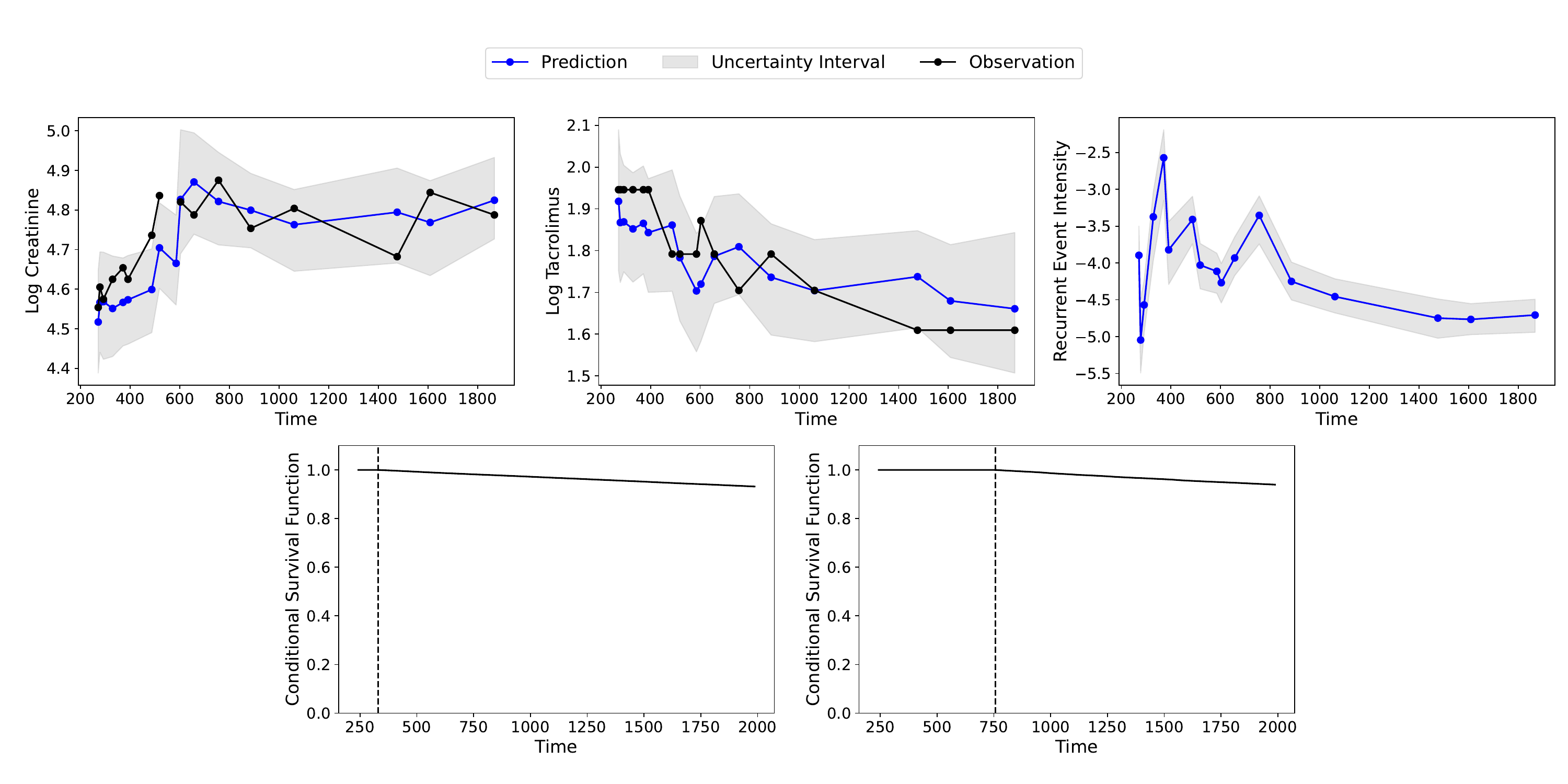}
        \caption{Patient S1, terminal event not observed.}
    \end{subfigure}%
    \newline 
    \begin{subfigure}{\textwidth}
        \centering
        \includegraphics[width=\linewidth]{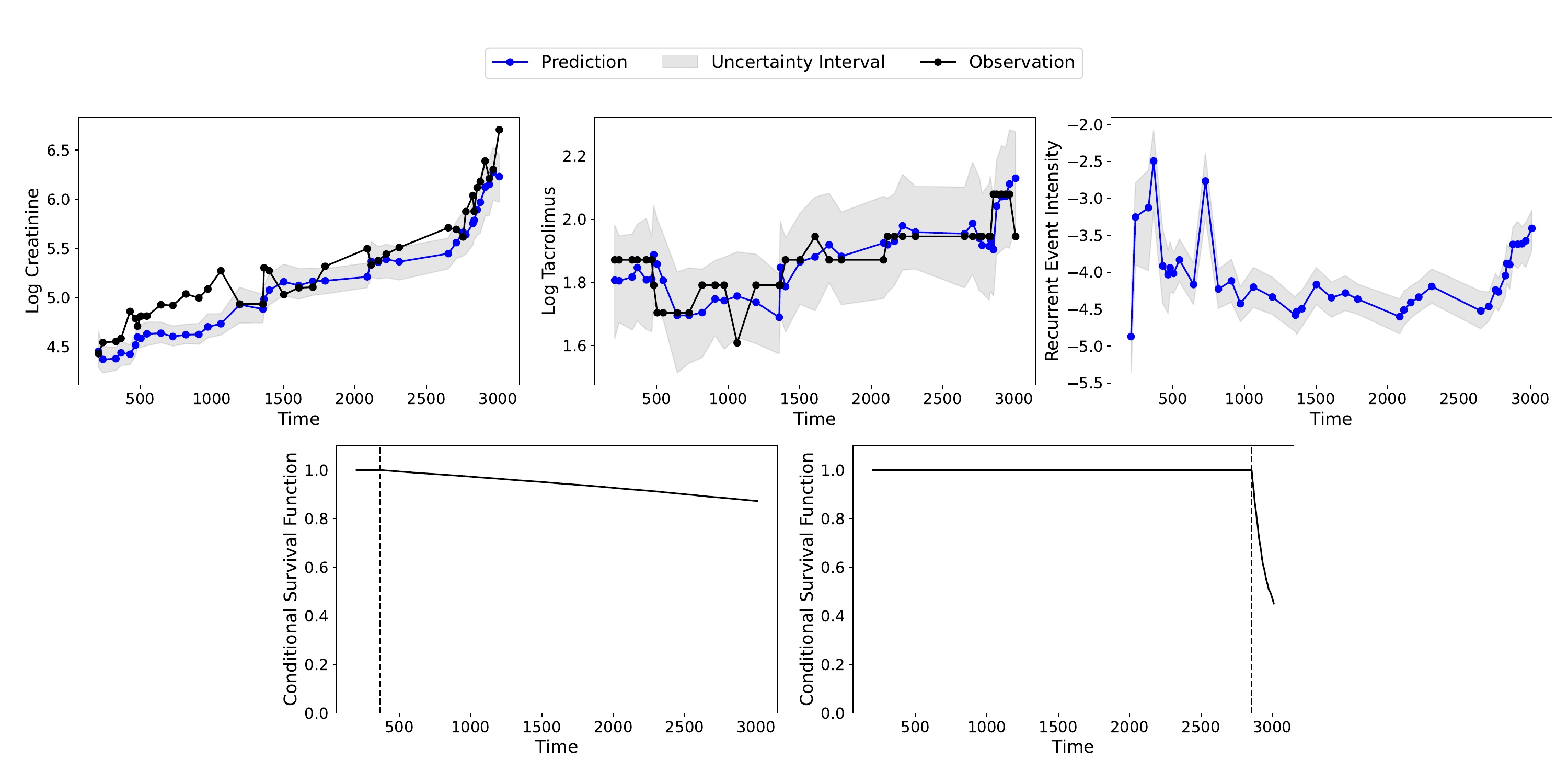}
        \caption{Patient S2, terminal event observed.}
    \end{subfigure}
    \caption{Dynamic prediction case studies for two patients from the DIVAT kidney transplantation dataset. Top panel: left, prediction of log creatinine level at the next visit vs. actual observation; middle, prediction of log tacrolimus dosage at the next vs. actual assignment; right, predicted log recurrent event intensity at each clinic visit. Bottom panel: left, conditional survival function given the history of the first five clinic visits; right, conditional survival function given the history up to the sixth last visit.}
    \label{fig: real_S1S2}
\end{figure}

 We then demonstrated the effectiveness of TransformerLSR in joint modeling of longitudinal data, survival, and recurrent events by performing dynamic prediction case studies on two individual patients: S1 and S2. Given the history $\mathcal{H}_j$, we dynamically predicted the log creatinine level at the next visit $j+1$, as well as the next assigned tacrolimus dosage. Additionally, at each visit $j$, we outputted the predicted recurrent event intensity $\lambda(t_j)$. To obtain uncertainty intervals for the longitudinal variables and intensity function, we employed the Monte Carlo dropout procedure outlined in \citep{gal2016theoretically}. Specifically, for each prediction, we generated $100$ Monte Carlo dropout samples, utilized the mean of the samples as the predicted value, and constructed the uncertainty interval using the $0.05$ and $0.95$ quantiles. To illustrate how TransformerLSR dynamically updated the survival hazard function as more information was incorporated into the patient history, we selected two landmark times for patients S1 and S2. By conditioning on the survival up to these landmark times as well as the corresponding history information, we computed the conditional survival function until each patient's survival event time or censoring time. The results for patients S1 and S2 are presented in \Cref{fig: real_S1S2}.

\par For patient S1, the trends of both the creatinine level and assigned tacrolimus dosage were accurately captured by TransformerLSR, as nearly all observations fell within the uncertainty intervals. The stability of patient S1's creatinine level and the drop in their assigned tacrolimus dosage both indicated effective functioning of S1's transplanted kidney. The recurrent event intensity also reflected the change in clinic visit schedule: as the visits became sparse towards the end, the intensity function dropped down correspondingly. A sparse visit schedule was also suggestive of the patient's stable health condition.
The conditional survival function at the early landmark time (around day 260) suggested that S1 was subject to very low survival risk, and with more information available, the trend of the updated survival function (around day 750) did not alter. Indeed, no death event was observed for S1 until the end of his record (around day 2000).

\par For patient S2, TransformerLSR also captured the general trends of the creatinine level and tacrolimus dosage. In contrast to patient S1, both the creatinine level and the assigned dosage of patient S2 gradually increased over time, indicating that S2 was subject to escalating risk as time progressed. The recurrent event intensity function also reflected the decrease in frequency of clinic visits since day 1000, followed by an increase in frequency towards the end. In terms of survival modeling, conditioning on the early history of S2 (around day 400), during which the creatinine level was low and the assigned dosage was stable, the conditional survival function decreased at a minimal rate. However, as more data became available, the conditional survival function at the second landmark time (around day 2850) decreased rapidly. This finding aligned with S2's medical record, as the death of patient S2 was observed shortly after the second landmark time (around day 3000).

\par
\begin{figure}
    \centering
    \quad
    \begin{subfigure}{\textwidth}
        \centering
        \includegraphics[width=\linewidth]{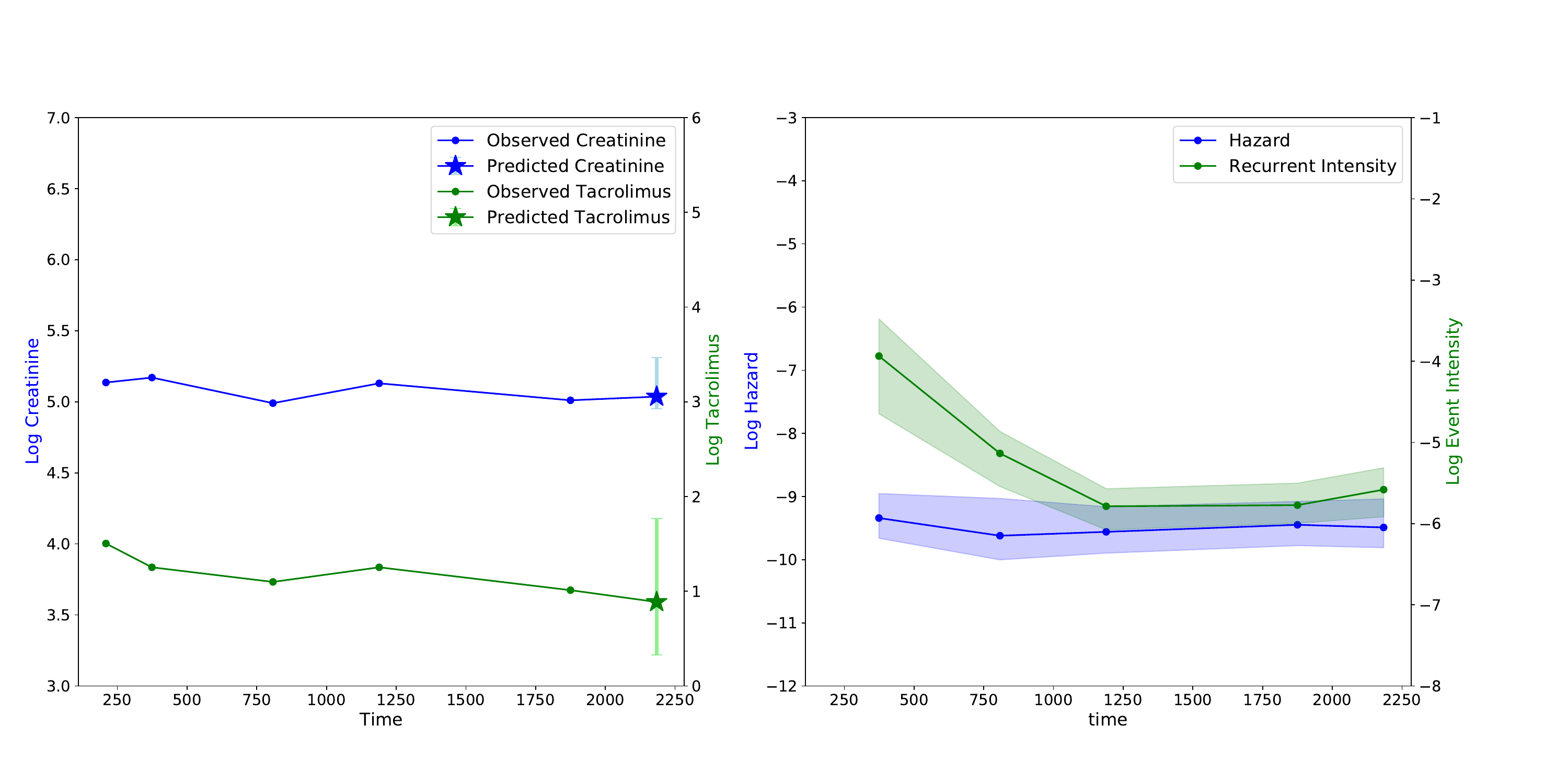}
        \caption{Patient S3, with predicted next visit interval time of 309 days.}
    \end{subfigure}%
    \newline 
    \begin{subfigure}{\textwidth}
        \centering
        \includegraphics[width=\linewidth]{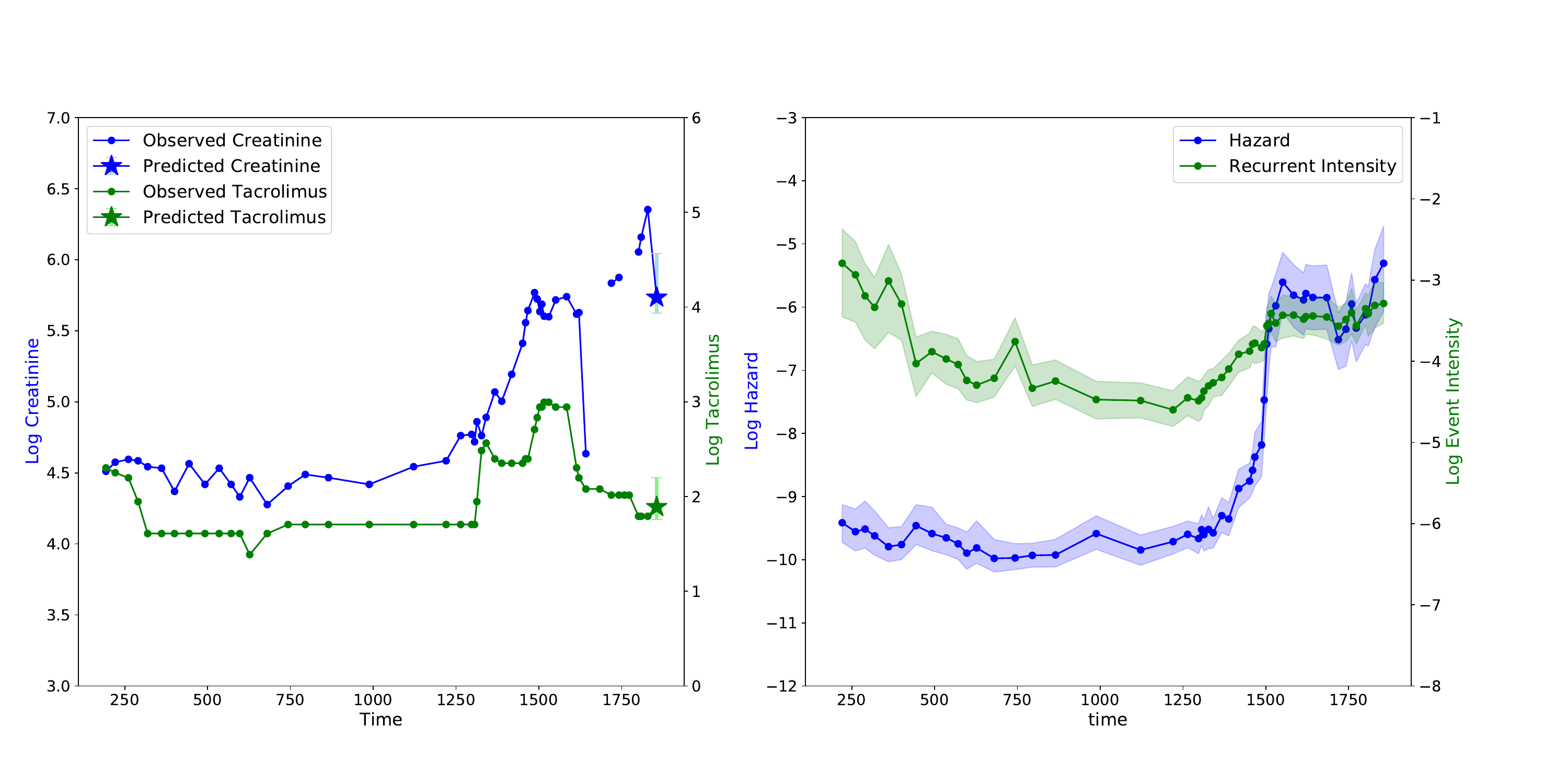}
        \caption{Patient S4, with predicted next visit interval time of 27 days.}
    \end{subfigure}
    \caption{Rollout case studies for two patients from the DIVAT kidney transplantation dataset. Left panel: predicted creatinine level and tacrolimus dosage based on their previous observed values. Right panel: fitted hazard function and recurrent event intensity values at the observed visit times and predicted next visit time.}
    \label{fig: real_S3S4}
\end{figure}

 By leveraging TransformerLSR's comprehensive understanding of the complex interactions between longitudinal measurements, survival probabilities, and recurrent events, healthcare providers can make informed decisions and deliver more tailored and effective patient care. 
To examine the capability of individualized predictions for TransformerLSR, we performed a one-step prognostication into the future based on the entire observed history of patients. Specifically, we randomly selected two patients, S3 and S4, from among all patients with right-censored records who survived up to their last clinic visit. We predicted their expected next clinic visit time, $t_{J_i+1}$, by sampling and averaging $100$ Monte Carlo visit times from the deep temporal point process with intensity $\lambda(t)$ estimated by TransformerLSR. Each sample was drawn using the thinning algorithm detailed in \ref{sec: append_thinning}. Then, conditioning on the estimated next visit time, $t_{J_i+1}$, we predicted the next creatinine level and assigned dosage. We also outputted the fitted hazard function value and recurrent event intensity value at each observed visit time, as well as the predicted next visit time. The uncertainty intervals of the longitudinal variables and intensity function values were again obtained using the Monte Carlo dropout procedure. The one-step rollout profiles of patients S3 and S4 are summarized in \Cref{fig: real_S3S4}.

\par The observed medical history of patient S3 exhibited a steady trend in creatinine levels and tacrolimus dosages over time. Given the stable health condition of S3, the predicted next visit interval time, defined as $t_{J_i+1}-t_{J_i}$, was 309 days for S3, suggesting that only regular annual follow-ups were necessary. The forecasted tacrolimus dosage for S3 showed a minor downward adjustment, while the creatinine level was expected to remain stable. Correspondingly, the projected hazard function for S3 exhibited negligible variation for the forthcoming visit. In contrast, patient S4's expected interval was much shorter, at 27 days, reflective of their increased frequency of clinic visits observed towards the latter part of their medical history. This aligned with the observed escalation in creatinine levels for S4, indicating a deteriorating health condition that demanded increased medical intervention. The anticipated creatinine level for S4 was projected to remain elevated, with the hazard function continuing its upward trend at the time of the predicted next visit. The marked disparity in predictions for the next visit intervals, alongside the divergent trajectory predictions of the longitudinal variables for patients S3 and S4, underscored the capability of TransformerLSR to provide personalized predictions that considered the comprehensive history of each patient.

In summary, the numerical results and case studies presented illustrate the robust capabilities of TransformerLSR in simultaneously modeling longitudinal data, survival outcomes, and recurrent events while effectively capturing the interdependence among these variables. Beyond its modeling capacity, TransformerLSR holds promise for clinical utility by providing personalized medical recommendations tailored to each patient's unique disease progression and medical history. For instance,  TransformerLSR  can offer valuable insights into the optimal timing for the next clinic visit, taking into account individual patient characteristics and evolving health conditions.

\section{Conclusion}
\label{sec: conclusion}

In this paper, we introduce TransformerLSR, the first deep attentive joint model of longitudinal data, survival, and recurrent events. TransformerLSR utilizes the powerful transformer architecture to capture the interdependence among the three components. Unlike prior deep joint models of longitudinal data and survival events, which typically model concurrent longitudinal measurements as multivariate variables, TransformerLSR introduces a novel trajectory representation. This representation enables each concurrent variable to be predicted autoregressively, with the sequence order of the variables refined by known clinical knowledge.
Furthermore, TransformerLSR models both recurrent events and survival events using deep point processes, without making assumptions about the parametric form of the intensity functions as in traditional statistical literature. Through extensive simulation studies, TransformerLSR demonstrates superior performance in predicting longitudinal variables and modeling survival compared to alternative methods. Additionally, it successfully recovers the simulation ground truth for recurrent event intensities. When applied to the DIVAT kidney transplantation dataset, TransformerLSR provides meaningful inference and interpretable results, further highlighting its effectiveness and potential utility for application in clinical settings.

The deep architecture of TransformerLSR offers high scalability and flexibility, making it suitable for a variety of biomedical studies and epidemiology applications that involve longitudinal data, survival events, and recurrent events, beyond the kidney transplantation study explored in this paper. Building on the framework of TransformerLSR, we identify the following promising future research directions. First, while our trajectory representation already hints at enhanced interpretability since it assumes sequential dependence among the concurrent variables, one can apply the technique mentioned in \citep{chefer2021transformer} to TransformerLSR's architecture to further explore such interpretability beyond attention weights. Second, we can further separate treatments from longitudinal variables modeling to incorporate elements from Causal Transformer \citep{melnychuk2022causal} and extend TransformerLSR to estimate counterfactual outcomes over time \citep{robins2008estimation}.


\appendix

\section{Asynchronous missing data}
\label{sec: append_missing}
Here, we elaborate on the way to handle asynchronous missing data (some but not necessarily all longitudinal variables may be missing at some observation times) without imputation. Consider the simple case where we aim to utilize the longitudinal observations $\vec{Y}_1 = (Y_{1,1},Y_{1,2},Y_{1,3})$ observed at $t_1$, and $\vec{Y}_2 = (Y_{2,1},Y_{2,2},Y_{2,3})$ observed at $t_2$ to make forecasts at $t_3$, and $Y_{1,2}$, the measurement for the second variable at $t_1$ is missing. The trajectory representation for TransformerLSR encoder input is $(Y_{1,1},\times, Y_{1,3},Y_{2,1},Y_{2,2},Y_{2,3})$, where $\times$ indicates that $Y_{1,2}$ is masked off. As such, the observed information $Y_{1,1}$ and $Y_{1,3}$ are still fully utilized. In the context of statistical modeling, we may think of the predictions $h(t_3|\mathcal{H}_2), \lambda(t_3|\mathcal{H}_2)$, and $\widehat{\vec{Y}}(t_3|\mathcal{H}_2)$ as conditional distributions with the missing variable $Y_{1,2}$ integrated out from the incomplete history $\mathcal{H}_2$. As demonstrated in \citep{mei2017neural} as well as our simulation studies in Section 3, such implicit marginalization procedure with deep models tends to perform well in practice. 

\par In comparison, TransformerJM, which also adopts the transformer architecture, masks off the entire $\vec{Y}_1$. This
results in a trajectory representation $(\times,\vec{Y}_2)$, completely ignoring the available information $Y_{1,1}$ and $Y_{1,3}$ which are taken out together with $Y_{1,2}$. Further, since TransformerJM considers a discrete time survival formulation, its conditional probability prediction is much more ``aware" of the relative position of the prediction token (e.g., the prediction token is the second token in the trajectory, with only $\vec{Y}_2)$ preceding it since $\vec{Y}_1$ is masked off), rather than the absolute time of the prediction token (e.g., at time $t_3$). As such, masking off entire visits becomes even more detrimental for TransformerJM survival prediction beyond having less information in the history.

\section{Likelihood functions}
\label{sec: append_like}
The log-likelihood function for $J$ recurrent events at times $\{t_j\}_{j=1}^J$ over the time interval $[0,T)$, as given in Section 2, is

\begin{equation}
    l_{\lambda} = \underbrace{\sum_{j=1}^{J} \log \lambda(t_j)}_{\text{event log-likelihood}} \qquad - \underbrace{\int_{0}^T \lambda(t) dt}_{:=\Lambda, \text{ non-event log-likelihood}} \! \! \! \! \!\!.
    \label{eq: sup_event_loglik}
\end{equation}

The proof to \Cref{eq: sup_event_loglik} may be found in standard point processes texts such as \citep{illian2008statistical}. We provide a proof here for completeness. 

\begin{proof}
The likelihood function of the $J$ events can be factored as the product of the conditional intensity functions of each event given its prior history:

\begin{equation}
L = \prod _{j=1}^Jf(t_j|\mathcal{H}_{j-1})(1-F(T|\mathcal{H}_{J})),
\label{eq: sup_lik_factor}
\end{equation}

where the last term $(1-F(T|\mathcal{H}_{J}))$ corresponds to the fact that no event happens over the interval $[t_J,T)$. Recall the definition of the conditional density $\lambda(t)$ for any $t>t_j$:
\begin{equation*}
    \lambda(t)=\lim_{dt\rightarrow 0} \frac{\mathbb{P}\{t\leq t_{j+1} <t+dt|t_{j+1}\geq t,\mathcal{H}_{j}\}}{dt},
\end{equation*}
by expanding the conditional probability in the numerator as the ratio between $\mathbb{P}\{t\leq t_{j+1} <t+dt|\mathcal{H}_{j}\}=f(t|\mathcal{H}_j)dt$ for infinitesimal $dt$ and $S(t|\mathcal{H}_j)=1-F(t|\mathcal{H}_j)$, we obtain: 
\begin{equation}
\lambda(t)= \frac{f(t|\mathcal{H}_{j})}{1-F(t|\mathcal{H}_{j})}.
\label{eq: inten}
\end{equation}

Further, from \Cref{eq: inten} we have
\begin{equation*}
\lambda(t) = \frac{\frac{d}{dt}F(t|\mathcal{H}_{j})}{1-F(t|\mathcal{H}_{j})} = -\frac{d}{dt}\log (1-F(t|\mathcal{H}_{j})).
\end{equation*}
Integrating both sides from $t_j$ to $t$, we get $\int_{t_j}^t \lambda(x) dx = -\log(1-F(t|\mathcal{H}_j))$, since events do not overlap, i.e., $F(t_j|\mathcal{H}_j) = 0$. Differentiating $F(t|\mathcal{H}_{j})$ with respect to $t$, we obtain

\begin{equation}
f(t|\mathcal{H}_{j}) = \lambda(t) \exp \left(-\int_{t_j}^t\lambda(x) dx \right).
\label{eq: CDF}
\end{equation}
Plugging \Cref{eq: inten} and \Cref{eq: CDF} into \Cref{eq: sup_lik_factor}:
\begin{align*}
L &= \left(\prod_{j=1}^J \lambda(t_j)\exp \left(-\int_{t_{j-1}}^{t_j} \lambda(x)dx\right)\right)\exp \left(-\int_{t_{J}}^{T} \lambda(x)dx\right)        \\
& = \left(\prod_{j=1}^J \lambda(t_j)\right)\exp \left(-\int_{0}^{T} \lambda(x)dx\right).
\end{align*}
Taking log of both sides, we obtain the log-likelihood function in \Cref{eq: sup_event_loglik}.
\end{proof}

For the survival process where there is at most $1$ event that is subject to right censoring, the summation in \Cref{eq: sup_event_loglik} is replaced by a singleton element evaluated at the survival/censoring time and becomes
\begin{equation}
l_h = (1-\delta) \log h(T) - \underbrace{\int_0^T h(t)dt}_{:=\zeta},
\label{eq: sup_surv_loglik}
\end{equation}
where $\delta=\mathbb{I}_{(C\leq E )}$ is the administrative censoring time.

\section{Thinning algorithm}
\label{sec: append_thinning}
The thinning algorithm we use is based on Lewis and Shedler, 1979, Algorithm 1 \citep{lewis1979simulation}. Let $t_0$ denote the current visit time and $T$ denote an upper bound for the next clinic visit time, \Cref{alg: thin} describes the procedure we use for the sampling of the next visit time. Since the recurrent event intensity $\lambda(t)$ is parameterized by a deep transformer model, seeking the maximum value over the interval $(t_0,T]$ is infeasible theoretically, we resort to using Monte Carlo samples that lie uniformly in the interval  $(t_0,T]$ and take the maximum as $\lambda^*$, the maximum over the interval. Due to the parallel nature of our transformer architecture, computing the Monte Carlo samples $\lambda(t)$ for $t\in (t_0,T]$ requires only one forward pass of the model. Analogously, by utilizing batch computing methods, drawing multiple samples (to compute e.g., the \emph{expected} next time) with the thinning algorithm \Cref{alg: thin} requires also only one forward pass rather than loop computation for efficiency. 
\begin{algorithm}[H]    
\caption{Thinning algorithm to sample the next event time on $(t_0,T]$.}
\label{alg: thin}
\begin{algorithmic}
\STATE {\bfseries Input:} $\lambda(t)$ given by the model; current time $t_0$, boundary time $T$
\STATE Initialize $\lambda^* = \sup _{t_0 < t \leq T} \lambda(t)$, $s=s_0=t_0$, \texttt{accepted} = \texttt{False};
\WHILE{$s < T$}
\STATE sample $w \sim \text{Exponential}(\lambda^*)$;
\STATE set $s = s_0 + w$;
\STATE sample $u \sim \text{uniform}(0,1)$;
\IF{$u\leq \lambda(s)/\lambda^*$}
\STATE \texttt{accepted} = \texttt{True};
\STATE \textbf{Break};
\ENDIF
\STATE set $s_0 = s$;
\ENDWHILE

\IF{\texttt{accepted}}
\STATE {\bfseries Return $s$}
\ELSE
\STATE  {\bfseries Return $T$}
\ENDIF
\end{algorithmic}
\end{algorithm}



\section{Additional plots for DIVAT case study}
\label{sec: append_divat}
\begin{figure}[ht!]
    \centering
    \includegraphics[width=0.85\linewidth]{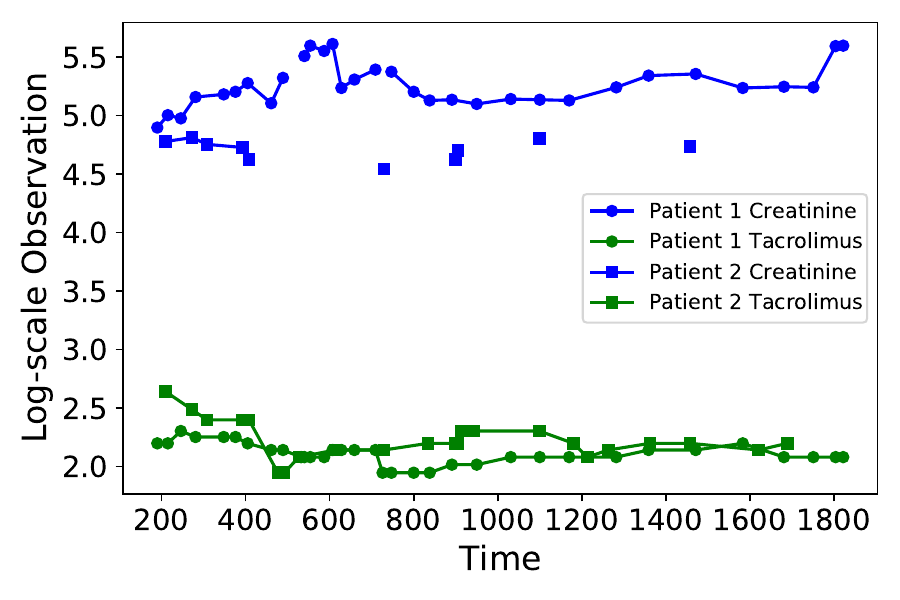}
        \caption{Data trajectory examples for two randomly selected patients, with each point representing a measurement at a clinic visit. Missingness in data is shown by disconnected lines.}
    \label{fig: data}
\end{figure}

We plot the log-scale creatinine and tacrolimus levels for two randomly selected patients to showcase the DIVAT dataset in \Cref{fig: data}.

\par
\begin{figure}[ht!]
    \centering
    \includegraphics[width=0.85\linewidth]{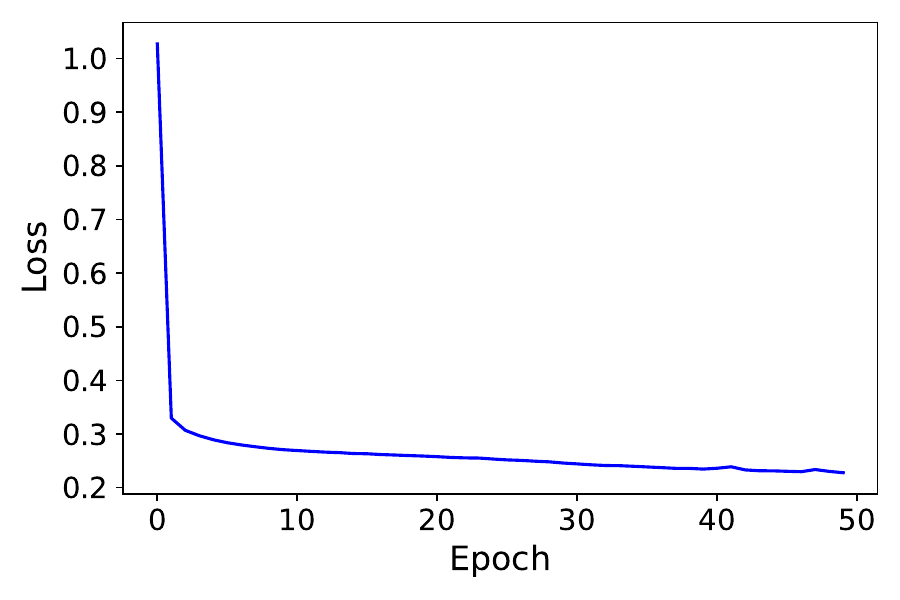}
        \caption{Training loss vs epoch.}
    \label{fig: train}
\end{figure}

 To show that 50 epochs was sufficient for training on the DIVAT dataset, we plot the training loss as a function of training epoch in \Cref{fig: train}.

\pagebreak
\newpage



\section*{Funding}
This work was supported by the National Science Foundation, USA [grant numbers 1940107, 1918854] and the National Institute of Health, USA [grant number R01MH128085]. 

\section*{Role of the funding source}
The study sponsor had no involvement in this manuscript.

 \bibliographystyle{elsarticle-num} 
    \bibliography{ctjm}
\end{document}